\documentclass{article}

\usepackage{float}
\usepackage{wrapfig}

\usepackage[final,nonatbib]{neurips_2024}

\usepackage[utf8]{inputenc} 
\usepackage[T1]{fontenc}    
\usepackage{hyperref}       
\usepackage{url}            
\usepackage{booktabs}       
\usepackage{amsfonts}       
\usepackage{nicefrac}       
\usepackage{microtype}      
\usepackage{xcolor}         
\usepackage{threeparttable}
\usepackage{multirow}
\usepackage{multicol}
\usepackage{subfigure}
\usepackage{amsfonts}
\usepackage{bm}
\usepackage{booktabs}
\usepackage{amsmath}
\usepackage{graphicx} 
\usepackage{caption} 

\title{CausalStock: Deep End-to-end Causal Discovery \\ for News-driven Stock Movement Prediction}

\author{%
Shuqi~Li$^{1}$\thanks{The first two authors contributed equally.}
  \quad Yuebo~Sun$^{1}$\footnotemark[1] \quad Yuxin Lin$^{2}$ \quad
\textbf{Xin~Gao}$^{3}$\quad \textbf{Shuo~Shang}$^{4}$ \quad \textbf{Rui~Yan}$^{1}$\thanks{Corresponding author: Rui Yan (\url{ruiyan@ruc.edu.cn}).} \\
  $^{1}$ Gaoling School of Artificial Intelligence, Renmin University of China\\
  $^{2}$ Peking University
  $^{3}$ King Abdullah University of Science and Technology\\
  $^{4}$ University of Electronic Science and Technology of China \\
  \texttt{\{shuqili, sunyuebo0418, ruiyan\}@ruc.edu.cn} \\
\texttt{linyuxin@stu.pku.edu.cn}, \texttt{xin.gao@kaust.edu.sa}, \texttt{jedi.shang@gmail.com}\\
}

\begin{document}

\maketitle

\begin{abstract}
  There are two issues in news-driven multi-stock movement prediction tasks that are not well solved in the existing works. On the one hand, ``relation discovery'' is a pivotal part when leveraging the price information of other stocks to achieve accurate stock movement prediction. Given that stock relations are often unidirectional, such as the ``supplier-consumer'' relationship, causal relations are more appropriate to capture the impact between stocks. On the other hand, there is substantial noise existing in the news data leading to extracting effective information with difficulty. With these two issues in mind, we propose a novel framework called CausalStock for news-driven multi-stock movement prediction, which discovers the temporal causal relations between stocks. We design a lag-dependent temporal causal discovery mechanism to model the temporal causal graph distribution. Then a Functional Causal Model is employed to encapsulate the discovered causal relations and predict the stock movements. Additionally, we propose a Denoised News Encoder by taking advantage of the excellent text evaluation ability of large language models (LLMs) to extract useful information from massive news data. The experiment results show that CausalStock outperforms the strong baselines for both news-driven multi-stock movement prediction and multi-stock movement prediction tasks on six real-world datasets collected from the US, China, Japan, and UK markets. Moreover, getting benefit from the causal relations, CausalStock could offer a clear prediction mechanism with good explainability.
  
\end{abstract}

\section{Introduction}
\label{sec:intro}

The financial services industry has maintained a leading position in embracing data science methodologies to inform investment determinations. Within this domain, quantitative trading has garnered substantial attention from both academia and industry. Researchers have consistently worked on exploring different approaches to predict the stock movement (rise or fall of stock price) for many years, such as uni-stock movement prediction~\cite{li2023pen}, multi-stock movement prediction~\cite{yoo2021accurate, luo2023causality}, news-driven stock movement prediction~\cite{xu-cohen-2018-stock, kim2019hats} and so on, which have shown significant success. These methods usually model the stock movement prediction task as a time series classification problem. 

In this paper, we focus on the news-driven multi-stock movement prediction task. A prevalent model paradigm for this task often takes the historical price features and the stock-related news of multiple stocks as inputs and then leverages the well-designed neural networks to make stock movement predictions. There are two key modeling points for tackling this task: modeling the stock relations to enhance the prediction accuracy, and building the text mining module to extract effective information from news data that benefits stock movement prediction. Although previous work has made significant progress, there are still some issues that require further attention. We will elaborate on them in the following.

For stock relation modeling, many existing works are commonly attention-based~\cite{hu2018listening, kim2019hats, luo2023causality} or graph-based~\cite{shih2019temporal, luo2023causality}. These methods aim to model the correlation relation between stocks. However, the company relations are often unidirectional,
such as the ``investing'' and ``member of,'' leading to the unidirectional relations of their stocks. Thus, causal relations are
more appropriate for depicting the impact between stocks, as they identify the direction of information flow and are more informative than correlations. With the development of causal science, many researchers have started to use deep end-to-end networks for
causal relations discovery of panel data or temporal data ~\cite{geffner2022deep,gong2022rhino}, in which the causal relations are defined as directed acyclic graphs, i.e., causal graphs, and the Functional Causal Models (FCMs) are often utilized to optimize the causal graph by simulating the data generation mechanism. This provides a solid theoretical foundation for causal discovery for stocks. 

In recent years, an extrinsic text mining module has emerged as a plausible avenue through the alignment of financial news and social media posts, thereby elucidating intricate market insights that extend well beyond mere considerations of price dynamics, trading volumes, or financial indicators~\cite{xing2018natural,jiang2021applications,si2013exploiting,schumaker2009textual}. Conventional text representations obtained by using GRU~\cite{xu-cohen-2018-stock} or LSTM~\cite{hu2018listening} exhibit many limitations. Specifically, news text data are often characterized by substantial noise because of the presence of irrelevant or ambiguous information~\cite{0News,2001News,2014News}. The effective information for stock movement prediction gets intertwined with this noise, presenting a considerable challenge for these modules to discern meaningful signals accurately. In contrast, Large Language Models (LLMs) have unique advantages in this situation due to their advanced knowledge and reasoning abilities. Besides, LLMs can identify meaningful information within noisy environments~\cite{roy2016solving, bubeck2023sparks}.

Motivated by these requirements, we propose an innovative news-driven multi-stock movement prediction model named CausalStock. In CausalStock, we design a Denoised News Encoder, which leverages LLMs to score every news text from multiple perspectives. Then the evaluation scores are taken as denoised text representations. To discover the causal relations between stocks, we propose a Lag-dependent temporal causal discovery module, from which we obtain the causal graph distribution.
Based on the input market information and learned causal graph distribution, CausalStock employs an FCM~\cite{gong2022rhino} to make predictions. We summarize the contributions of our paper as follows:

\begin{itemize}
    \item  We propose a novel news-driven multi-stock movement prediction method named CausalStock, which could discover the causal relations among stocks and make accurate movement predictions simultaneously.

    \item Different from the past lag-independent causal discovery method~\cite{geffner2022deep}, CausalStock involves a lag-dependent temporal causal discovery module, which intuitively links the temporal causal relations according to the time lag, making it more suitable for temporal stock data.  
    
    \item To extract useful information from the massive noisy news text data, an LLM-based Denoised News Encoder is proposed by taking advantage of the evaluation ability of LLM, which outputs the denoised news representation for better information utilization.

\end{itemize}

Experiments on $6$ public benchmarks show the performance of CausalStock as a news-driven multi-stock movement prediction method. Moreover, we conduct extensive analytical experiments to show the explainability of our key modules.

\section{Related work}
\label{app:related_work}
\paragraph{Stock prices prediction}

In traditional trading practices, there are two analysis paradigms commonly used to make stock movement predictions:  technical analysis and fundamental analysis. With technical analysis, investors and traders tend to forecast stock prices relying on historical price patterns. Fundamental analysis aims to assess the intrinsic value of a stock by considering other factors besides historical prices, such as financial statements, industry trends, and economic conditions. 

Since stock movement prediction involves sequential data, RNN-based networks are applied in many works. ALSTM ~\cite{qin2017dual} integrated a dual-stage attention mechanism with LSTM. Adv-ALSTM ~\cite{feng2018enhancing} further employed adversarial training by adding perturbations to simulate the stochastic and unstable nature of the price variable. In recent years, researchers also exploited attention-based mechanisms to model complex interactions. DTML ~\cite{yoo2021accurate} is proposed to predict by using a transformer and LSTM to capture the asymmetric and dynamic correlations between stocks. With the development and prosperity of NLP technology, text from social media and online news has become a new popular source of fundamental analysis. HAN ~\cite{hu2018listening} designed two attention networks to recognize both the influential time periods of a sequence and the important news at a given time. Stocknet ~\cite{xu-cohen-2018-stock} proposed a deep generative model with recurrent, continuous latent variables. MSHAN ~\cite{DBLP:conf/iconip/GongE21} exploited a multi-stage TCN-LSTM hybrid model.  PEN ~\cite{li2023pen} proposed a Shared Representation Learning module to capture interactions between price data and text data. Additionally, many works modeled the correlation between stocks to enhance stock price prediction. MAN-SF ~\cite{sawhney2020deep} constructed a graph attention network with price features, social media, and inter-stock relationships based on the interrelationship between price and tweets. CMIN ~\cite{luo2023causality} was proposed to model the asymmetric correlations between stocks by computing transfer entropy. In addition, Co-CPC ~\cite{wang2021coupling} modeled the dependence between a certain stock industry and relevant macroeconomic variables. All the aforementioned methods aim to discover the correlation relations among stocks, as elaborated before, the causal relations are more appropriate to depict the information flow of stocks. In this work, we aim to model the causal relations for better stock movement prediction performance.

\paragraph{Causal discovery}

The conventional approach to discovering causal relations typically involves conducting randomized experiments~\cite{pearl2009causal,glymour2019review}. 
However, conducting randomized experiments can often be excessively expensive, overly time-consuming, or impossible to execute. Consequently, causal discovery, which aims to infer causal relationships from purely observational data, has attracted considerable attention within the machine learning community over the last decade~\cite{cheng2023causaltime, glymour2019review}. Causal discovery can be classified into three groups: constraint-based~\cite{gerhardus2020high,runge2020discovering,runge2019detecting}, score-based~\cite{bellot2021neural,pamfil2020dynotears,zheng2018dags}, and functional causal models (FCMs)~\cite{glymour2019review, pamfil2020dynotears}. FCMs define the causal relations by directed acyclic graphs (DAGs) and identify causal links through nonlinear functions, such as neural networks~\cite{gong2022rhino, geffner2022deep, zheng2020learning, khemakhem2021causal}.
Specifically,  DECI~\cite{geffner2022deep} is a deep end-to-end framework to discover causal relations based on additive noise FCM.
After that, Rhino ~\cite{gong2022rhino} was proposed to tackle the temporal causal discovery problem, which incorporates non-linear relations, instantaneous effects, and flexible history-dependent noise. In this work, we focus on utilizing the FCM to discover stock relations.

\section{Preliminary \& problem formulation}
\subsection{Preliminary}

In CausalStock, we integrate the model inputs with causal relations into
FCM for prediction. In this section, we
introduce the fundamental concepts of FCM and the temporal causal graph.

\paragraph{Temporal causal graph}

Consider a multivariate time series $\{X_t^i\}_{i=1}^{D}$ with $D$ variables,
the temporal causal graph $\bm{G}$~\cite{zheng2018dags} is commonly defined as a series of directed acyclic graph $\bm{G} =\left[G_1, G_2, \dots, G_{L}\right]= \{G_{l}\}_{l=1}^{L} \in \mathbb{R}^{L\times D\times D}$ with maximum time lag $L$. Each $G_l \in \mathbb{R}^{D\times D}$ specifies the lagged causal relationships between $X_{t-l}$ and $X_t$, the element $G_{l,ji}=1$  if there exists a causal link $X_{t-l}^j \xrightarrow{} X_{t}^i$ and $G_{l,ji}=0$ otherwise.

\paragraph{Functional causal model (FCM)}

FCM represents a set of generative functions that incorporate the input features based on causal knowledge (structured as a causal graph) to produce a final prediction. Optimizing the prediction accuracy concurrently refines the underlying causal graph. The theoretical demonstration presented in ~\cite{gong2022rhino, geffner2022deep} indicates that if the prediction is accurate, the causal graph can be considered a reliable approximation of real causal relations. Given the temporal causal graph $\bm{G}$ defined as before, a temporal FCM is defined as follows:

\vspace{-1em} 
{\small
\begin{equation}
\label{eq:sem}
X^i_t=F_{i}\left(\mathbf{P a}^i_{\bm{G}}\left(<t\right), z^i_t\right),
\end{equation}}
\vspace{-1em}

where $\mathbf{P a}^i_{\bm{G}}\left(<t\right)$ indicates the time-lagged parent nodes of variable $X^i_t$ following the temporal causal graph $\bm{G}$ and $z^i_t$ represents mutually and serially independent exogenous noise. Here $F_{i}$ is a function which implies how variable $X^i_t$ depends on its parents and the noise $z^i_t$. Given the distribution of noises for different variables  $\{z_t^i\}_{i=1}^{D}$ and causal graph, this FCM induces a joint distribution of the multivariate time series process $\{X_t^i\}_{i=1}^{D}$.

\begin{wrapfigure}{R}{0.35\linewidth}
\centering
\vspace{-1.6cm}
\includegraphics[width=4.5cm]{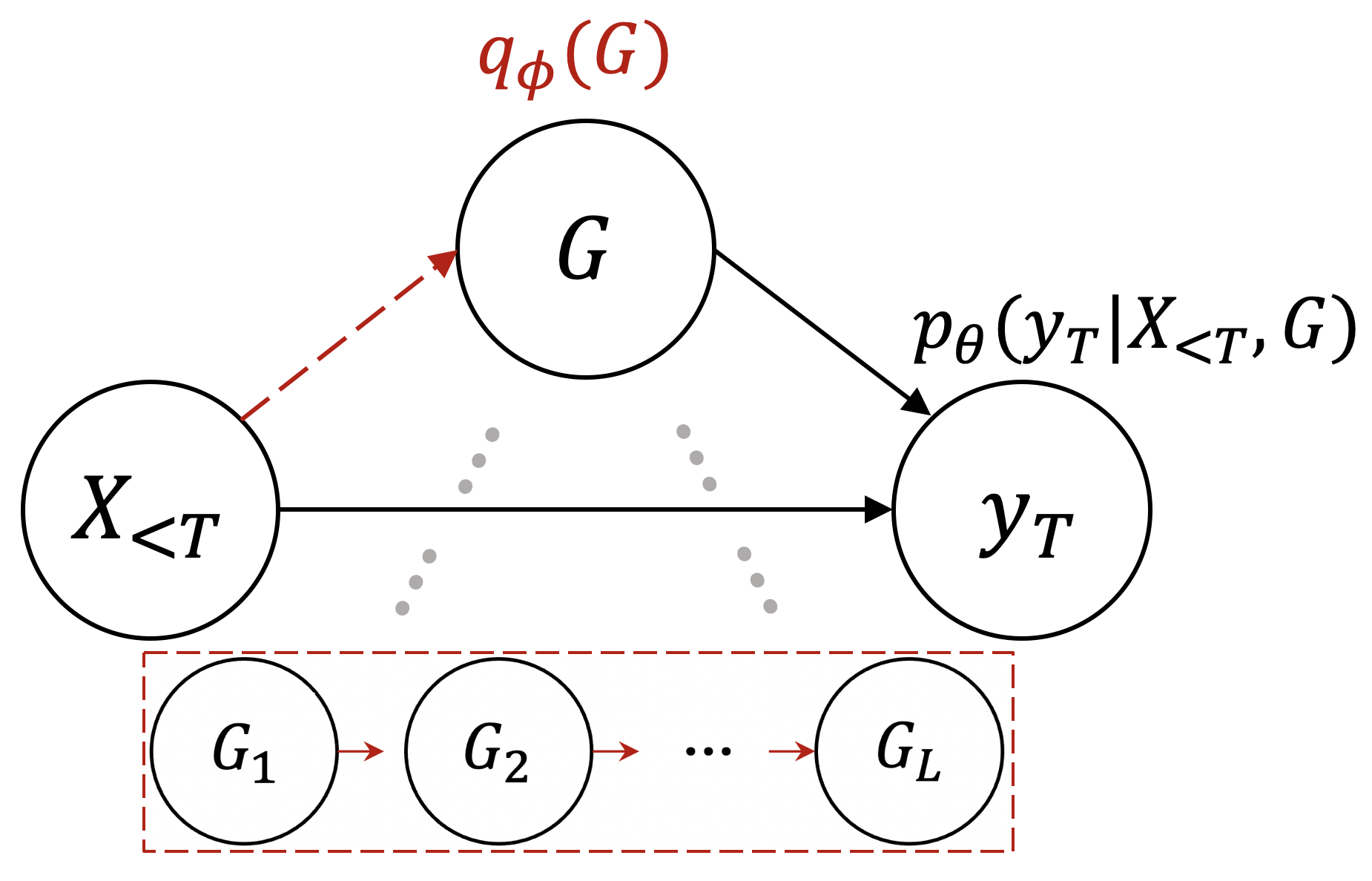}
\caption{Illustration of the process of stock movement $\bm{y}_T$ forecasting. The forecasting process is denoted by solid lines with parameters $\theta$ and the causal discovery process is denoted by dashed lines with variational approximation parameters $\phi$, $q_\phi$ is the posterior distribution of the causal graph.}
\label{fig:frame}
\vspace{-1cm}
\end{wrapfigure}

\subsection{Problem formulation}

In this paper, we focus on tackling the news-driven multi-stock movement prediction task. For the target trading day $T$, we denote the model inputs as the past $L$ time lag information of $D$ stocks as $\bm{X}_{<T} = \{X_t^i\}_{t=T-L:T-1}^{i=1:D} = [\bm{C}_{<T}, \bm{P}_{<T}]=\{[C_t^i, P_t^i]\}_{t=T-L:T-1}^{i=1:D}$, where $C_t^i$ and $P_t^i$ represent the news corpora representation and the historical price features representation of $i$-th stock at time step $t$ respectively. The objective is to predict the movement of adjusted close prices $\bm{y}_T=\{y_T^i\}_{i=1}^D \in 
\mathbb{R}^{D \times 1}$ on $T$-th trading day of all stocks simultaneously, where $y_T^i \in \{0,1\}$ representing the $i$-th stock price will fall or rise at trading day $T$, i.e., stock movement. In a theoretical way, this task could be trained by maximizing the log-likelihood of conditional probability distribution $ p \left(\bm{y}_T \mid \bm{X}_{<T}\right)$, so that the most likely $\bm{y}_T$ are generated.

\section{CausalStock}
\subsection{Model overview}

The conditional probability distribution could be further factorized as follows:

\vspace{-1em} 
{\small
\begin{equation} 
\begin{aligned}
\label{eq:model}
    p\left(\bm{y}_T \mid \bm{X}_{<T}\right) = \int_{\bm{G}} p\left(\bm{y}_T, \bm{G} \mid \bm{X}_{<T}\right) d\bm{G} = \int_{\bm{G}} p\left(\bm{y}_T \mid \bm{X}_{<T}, \bm{G}\right)p\left(\bm{G} \mid \bm{X}_{<T}\right) d\bm{G}.
\end{aligned}
\end{equation}}
\vspace{-1em}

The overall process is taken as two joint training parts: temporal causal graph discovery $p\left(\bm{G} \mid \bm{X}_{<T}\right)$ and the prediction process given the causal relations $p\left(\bm{y}_T \mid \bm{X}_{<T}, \bm{G}\right)$. The probabilistic graphic representation of this modeling process is shown in Figure~\ref{fig:frame}. In CausalStock, we develop a lag-dependent causal discovery module, according to which we could take another step by modeling $p\left(\bm{G} \mid \bm{X}_{<T}\right)$ as a lag-dependent format:

\vspace{-1em}
{\small
\begin{equation}
\begin{aligned}
\label{eq:G}
p\left(\bm{G} \mid \bm{X}_{<T}\right)=p\left(G_{1} \mid X_{T-1}\right)\prod_{l=2}^{L}p\left(G_{l} \mid G_{l-1},X_{T-l}\right).
\end{aligned}
\end{equation}}
\vspace{-1em} 

For the prediction part $p\left(\bm{y}_T \mid \bm{X}_{<T}, \bm{G}\right)$, we design an FCM as shown in Equation~\ref{eq:anm} to predict the future movement based on the past information $\bm{X}_{<T}$ and the discovered temporal causal graph $\bm{G}$.

\begin{figure*}
  \includegraphics[width=\linewidth]{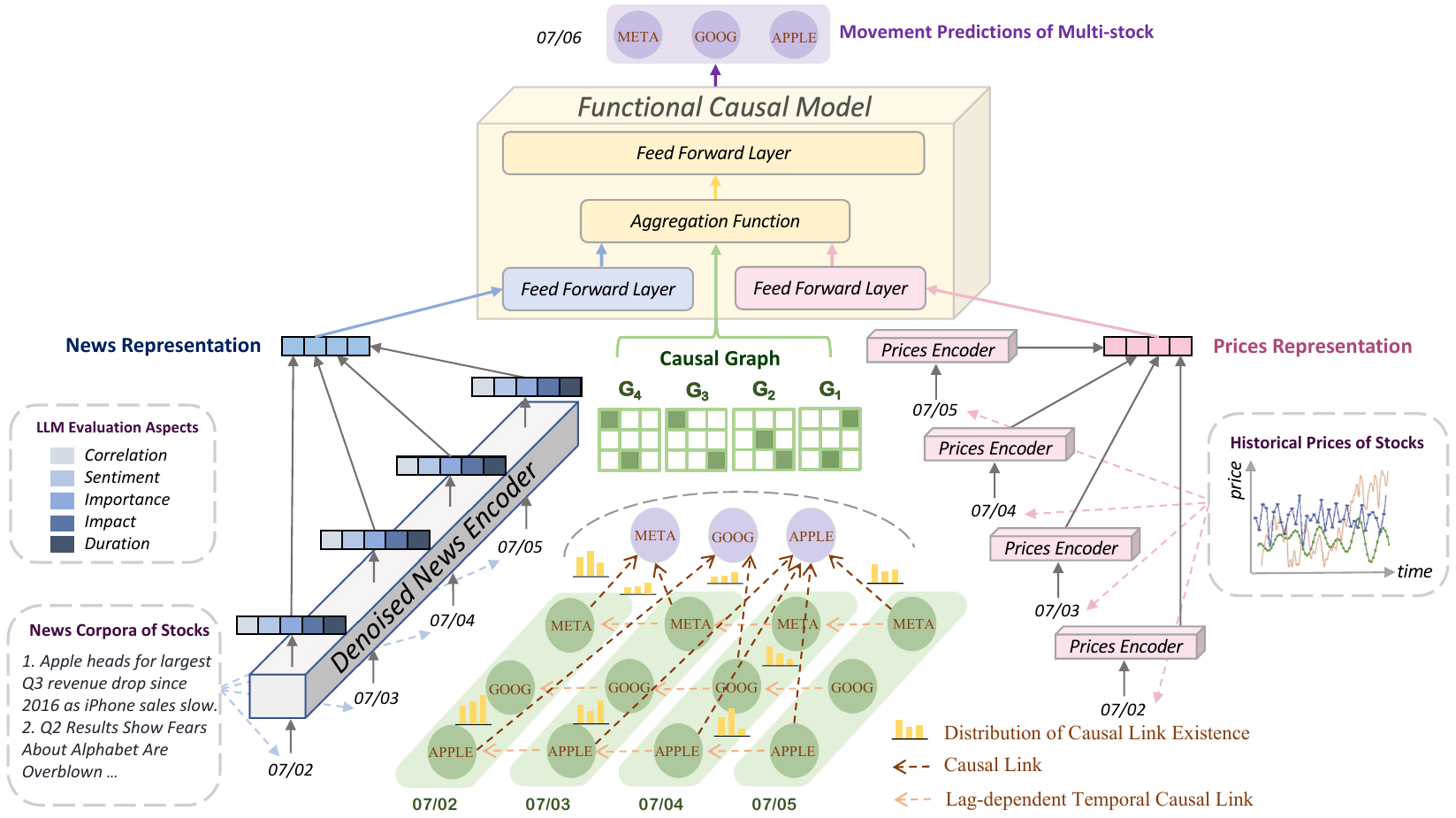}
  \caption{The structure of CausalStock. For illustration, we use market information during 07/02 - 07/05 of three stocks (AAPL, GOOG, META) to predict the movements of 07/05.}
  \label{fig:model}
\end{figure*}

In a nutshell, 
CausalStock comprises three primary components as shown in Figure~\ref{fig:model}:

\begin{enumerate}

\item Market Information Encoder (MIE) encodes the news text and price features. In this part, an LLM-based Denoised News Encoder is proposed;

\item Lag-dependent Temporal Causal Discovery (Lag-dependent TCD) module leverages variational inference to mine the causal relationship based on the given market information of stocks, i.e., modeling $p\left(\bm{G} \mid \bm{X}_{<T}\right)$;

\item Functional Causal Model (FCM) generates the prediction of future price movements according to the discovered causal graph, i.e., modeling $p\left(\bm{y}_T \mid \bm{X}_{<T}, \bm{G}\right)$.
\end{enumerate}

\subsection{Market information encoder (MIE)}

Market Information Encoder (MIE) takes news corpora and numerical stock price features as inputs, and outputs the historical market information representations $\bm{X}_{<T} = [\bm{C}_{<T}, \bm{P}_{<T}]= \{[C_t^i, P_t^i]\}_{t=T-L:T-1}^{i=1:D} = \{X_t^i\}_{t=T-L:T-1}^{i=1:D}$ for $D$ stocks with time lag $L$. For $i$-th stock, each time step representation $X_{t}^i$ is the combination of the text representation $C_{t}^i$ generated by the news encoder and the historical price features representation $P^i_{t}$ generated by the price encoder.

\paragraph{Price encoder}

For $i$-th stock, we denote the raw adjusted closing, highest, lowest, open, closing prices and trading volume on trading day $t$ as
$\hat{P}^i_{t}=\left[\hat{P}_{t}^{i,a},\hat{P}_{t}^{i,h}, \hat{P}_{t}^{i,l}, \hat{P}_{t}^{i,o},\hat{P}_{t}^{i,c}, V_t\right]$. By feeding $\hat{P}^i_{t}$ into the embedding layer, the historical prices could be represented as $P^i_{t} \in \mathbb{R}^{d_p \times 1}$, where $d_p$ is the price embedding size.

\paragraph{LLM-based denoised news encoder (DNE)}

News Encoder aims to embed stock-related news text, which evolves from the small sequential module, e.g., GRU~\cite{luo2023causality}, to pre-trained models, e.g., Bert and Roberta~\cite{devlin2018bert,liu2019roberta}, offering greater performance and scalability. However, news text data often contains massive noise due to the following factors. Firstly, news comes from a wide range of sources with varying degrees of reliability and editorial standards. This variability contributes to inconsistencies and inaccuracies in the information presented. Secondly, the sheer volume of news content generated daily can lead to information overload, where significant information is buried under less relevant or redundant information. Thirdly, the use of complex or ambiguous language can also add noise, making it difficult to extract precise information relevant to specific needs, such as stock movement prediction. Addressing these challenges requires sophisticated text mining and natural language processing techniques to filter out noise and extract useful, accurate information from news text data. With the development of large language models, current large language models can accurately capture the meaning of text and have a strong capability to evaluate text. Therefore, here we propose an LLM-based Denoised News Encoder to tackle these standing challenges.

LLM-based Denoised News Encoder is an innovative textual representation approach that not only proficiently captures salient information from extensive news texts but also assimilates external knowledge derived from LLMs to enrich the representations. Specifically, we employ an LLM and devise a series of prompts (see Appendix~\ref{app:prompt} for the whole designed prompts) to analyze the relationship between a news text and a specific stock from five dimensions: correlation between the news and the stock, sentiment polarity of the news, significance of the news event, potential impact of the news on stock prices, and duration of the news impact. Each dimension is scored, with Correlation, Importance, Impact, and Duration ranging from $0$ to $10$, while Sentiment varies from $-1$ to $1$. Thus the $i$-th text at day $t$ is represented as a five-dimensional representation ${\hat{C}^i_{t}} \in \mathbb{R}^{l \times 5}$. After the embedding layer, we obtain the final denoised news embedding ${C^i_{t}} \in \mathbb{R}^{l \times d_m}$. This novel encoding method amalgamates information derived from the primary text, the external knowledge embedded and the evaluation ability within the LLM. Besides, this method effectively reduces the significant noise present in the original text data.

\subsection{Lag-dependent temporal causal discovery (Lag-dependent TCD)}

In this section, we propose Lag-dependent Temporal Causal Discovery module. Inspired by~\cite{gong2022rhino}, our model takes a Bayesian view for modeling the distribution of temporal causal graph, which aims to learn the posterior distribution $p\left(\bm{G} \mid \bm{X}_{<T}\right)$. Unfortunately, the exact graph posterior is intractable because of the large combination space of $\bm{G}$. Here we adopt the variational inference~\cite{blei2017variational} to get the approximator $q_\phi\left(\bm{G}\right)$, where $\phi$ indicates the parameter set of variational inference. 

\paragraph{Graph prior}

The prior $p\left(\bm{G}\right)$ consists of two parts: the graph sparseness prior and the domain-specific knowledge prior. The unnormalised graph prior is as follows,

\vspace{-1em} 
{\small
\begin{equation} 
p(\bm{G}) \propto \exp \left(-\lambda_s\left\|\bm{G}_{1: L}\right\|_F^2-\lambda_d\left\|\bm{G}_{1: L}- 
 \bm{G}_{1: L}^p\right\|_F^2\right),
\end{equation}}
\vspace{-1em}

where $\lambda_s$ and $\lambda_d$ are scalar weights of graph sparseness and domain-specific knowledge constraint; $\bm{G}^p$ is an optional domain-specific knowledge graph, which allows users to incorporate pre-defined knowledge for guiding CausalStock, turning it into a knowledge and data-driven framework. Suppose a sudden event affects the causal relationships between stocks, such as a company ending a partnership. By incorporating this new pre-defined knowledge into $\bm{G}^p$, the causal graph is dynamically updated to reflect the latest market structure and relationship changes. $\left\| \cdot \right\|_F$ denotes Frobenius norm. It should be noted that there is no need to give a DAG constraint for the temporal causal graph defined in our paper, it is DAG naturally for the irreversibility of time.

\paragraph{Variational approximating graph posterior}

According to Equation~\ref{eq:G}, we factorize the approximator $q_\phi\left(\bm{G}\right)$ in the same way. For each underlying causal link $G_{l,ji}$ in $\bm{G}$, we let the posterior $q_\phi\left(G_{l,ji} \mid G_{l-1,ji}\right)$ subject to a Bernoulli distribution $\bm{B}$. So that the probability distribution of $q_\phi\left(\bm{G} \right)$ could be a product of Bernoulli distributions as follows,

\vspace{-1em} 
{\small
\begin{equation}
\begin{aligned}
\label{eq:qG}
q_\phi\left(\bm{G}\right)=q_\phi\left(G_1\right)\prod_{l=2}^{L}q_\phi\left(G_l \mid G_{l-1}\right)=\prod_{i=1}^{D}\prod_{j=1}^{D}q_\phi\left(G_{1,ji}\right)\prod_{l=2}^{L}\prod_{i=1}^{D}\prod_{j=1}^{D}q_\phi\left(G_{l,ji} \mid G_{l-1,ji}\right).
\end{aligned}
\end{equation}}
\vspace{-1em} 
 
The existence and non-existence likelihood tensors of causal links are parameterized as $\bm{U}=\{U_{l}\}_{l=1}^{L}=\{u_{l,ji}\}_{l=1:L}^{j,i= 1:D} \in \mathbb{R}^{L \times D \times D}$ and $\bm{V} = \{V_{l}\}_{l=1}^{L}=\{v_{l,ji}\}_{l=1:L}^{j,i= 1:D} \in \mathbb{R}^{L \times D \times D}$ separately, where $u_{l,ji}$ indicates the likelihood for edge existence from $X_{T-l}^j$ to $y_T^i$ and $v_{l,ji}$ is the likelihood for no-edge, which are all learnable parameters. To model the dependency between $G_{l,ij}$ with $G_{l-1,ij}$, we propose the following transformation:

\vspace{-1em} 
{\small
\begin{equation}
u^{\prime}_{l,ji}=h_{u}\left(u_{l,ji}, u_{l-1,ji}\right), v^{\prime}_{l,ji}=h_{v}\left(v_{l,ji}, v_{l-1,ji}\right),
\label{eq:dependent_mechanism}
\end{equation}}
\vspace{-1em} 

where $h_u$ and $h_v$ are trainable 3-layer MLPs. After normalization, the link existence probability tensor is denoted as $\bm{\Sigma} = \{\bm{\Sigma}_{l}\}_{l=1}^{L}=\{\sigma_{l,ji}\}_{l=1:L}^{j,i= 1:D}$, 

\vspace{-1em} 
{\small
\begin{equation}
\sigma_{l, ji}=\exp \left(u^{\prime}_{l,ji}\right)/\left(\exp \left(u^{\prime}_{l,ji}\right)+\exp \left(v^{\prime}_{l,ji}\right)\right),
\end{equation}}
\vspace{-1em}

where $\sigma_{l,ji}$ represents the link probability from $X_{T-l}^j$ to $\bm{y}_T^i$. Thus, we could derive the variational posterior:

\vspace{-1em} 
{\small
\begin{equation}
\begin{aligned}
q_{\phi}\left(G_{1,ji}\right)\sim\mathbf{B}\left(1,\sigma_{1,ji}\right), q_{\phi}\left(G_{l,ji} \vert G_{l-1,ji}\right)\sim\mathbf{B}\left(1,\sigma_{l,ji}\right),
q_{\phi}\left(\mathbf{G}\right)\sim\prod_{l=1}^{L}\prod_{i=1}^{D}\prod_{i=1}^{D}\mathbf{B}\left(1,\sigma_{l,ji}\right).
\end{aligned}
\end{equation}}
\vspace{-1em}

In the training stage, we employ the Gumbel-softmax reparameterization~\cite{maddison2016concrete,jang2016categorical} to stochastically estimate the gradients with respect to $\phi$. Besides, we design another parameterized learnable causal weight graph ${\bm{\hat{G}} = \{\hat{G}_{l}\}_{l=1}^{L} \in \mathbb{R}^{L\times D\times D}}$ to measure the causal degree. The separate design of the causal existence graph and the causal weight graph allows for more comprehensive modeling of causality. Once our model is fitted, the time series causal graph $\bm{G}$ can be sampled by $\bm{G} \sim q_\phi\left(\bm{G}\right)$ to represent the relation network and information flow of the stock market.

\subsection{Functional causal model (FCM)}
\label{sec:prediction}
In this section, we design an FCM to model $p_{\theta} \left(\bm{y}_T \mid \bm{X}_{<T}, \bm{G}\right)$, where $\theta$ denotes the parameter set of FCM. We focus on additive noise FCM~\cite{khemakhem2021causal} to generate $\bm{y}_T=\{\bm{y}^i_T\}_{i=1}^D \in \mathbb{R}^{D\times 1}$:

\vspace{-1em}
{\small
\begin{equation}
\label{eq:anm}
\bm{y}^i_T=F_{i}\left( \mathbf{P a}^i_{\bm{G}}\left(<T\right), z_T^i\right)=f_{i}\left(\mathbf{P a}^i_{\bm{G}}\left(<T\right)\right)+z_T^i,
\end{equation}}
\vspace{-1em}

where $z^i_t$ represents mutually and serially independent dynamical noise, and $f_i: \mathbb{R}^{D \times L}\xrightarrow{}\mathbb{R}^1$ are general differentiable non-linear function that satisfies the relations specified by the temporal causal graph $\bm{G}$ strictly, namely, if $X_{t}^j \notin \mathbf{P a}_G^i(<T)$, then $\partial f_i / \partial X_{t}^j=0$.  

We design a novel FCM to aggregate market information including news and prices based on the discovered causal graph $\bm{G}$ and causal weight graph $\bm{\hat{G}}$:

\vspace{-1em}
{\small
\begin{equation}
\begin{aligned}
f_{i}\left(\mathbf{P a}^i_{\bm{G}}\left(<T\right)\right)=\text{Sigmoid}\left(\zeta_i\left(\sum_{l=1}^L \sum_{j=1}^D G_{l, ji}\hat{G}_{l, j i} \left[\ell\left(P_{T-l}^j\right), \psi\left(C_{T-l}^j\right)\right]\right)\right),
\end{aligned}
\end{equation}}
\vspace{-1em}

where $\zeta_i$, $\ell$ and $\psi$ are all neural networks. $\ell$ and $\psi$ are shared weights across nodes and lags for efficient computation. $[\cdot, \cdot]$ denotes the concatenate operation. We apply the logistic Sigmoid function to output the movement probability of $\bm{y}_T$ and use it directly as the output of CausalStock.

For the exogenous noise $\bm{z}_T^i$ modeling, we adopt Gaussian distribution, i.e., $z_T^i\sim \mathcal{N}\left(0, \left(\sigma^i\right)^2\right)$, where per-variable variances $\left(\sigma^i\right)^2, i \in \left[1, D\right]$ are trainable parameters to represent the uncertainty part. According to Change of variables formula~\cite{khemakhem2021causal}, the conditional distribution $p_\theta\left(y_T^i \mid\mathbf{P a}^i_{\bm{G}}\left(<t\right)\right)$ could be represented as:

\vspace{-1em}
{\small
\begin{equation}
\label{eq:X}
p_\theta\left(y^i_T \mid\mathbf{P a}^i_{\bm{G}}\left(<t\right)\right)=p_{z_i}\left(z_T^i \right)\left|\frac{\partial F_i}{\partial z_T^i}\right|^{-1}=p_{z_i}\left(z_T^i \right),
\end{equation}}
\vspace{-1em}

where $p_{z_i}$ is the aforementioned Gaussian distribution for stock $i$. $\left|\frac{\partial F_i}{\partial z_T^i}\right|$ indicates the absolute value of the Jacobian-determinant for $F_i$, $\left|\frac{\partial F_i}{\partial z_T^i}\right|^{-1}=1$ is derived according to Equation~\ref{eq:anm}. Now the log likelihood $\log p_\theta\left(\bm{y}_T \mid \bm{X}_{<T}, \bm{G}\right)$ could be further represented as:

\vspace{-1em}
{\small
\begin{equation}
\begin{aligned}
\log p_\theta\left(\bm{y}_T \mid \bm{X}_{<T}, \bm{G}\right)=\sum_{i=1}^D \log p_\theta\left(y_T^i \mid \mathbf{Pa}^i_{\bm{G}}(<T)\right)=\sum_{i=1}^D \log p_{z_i}\left(z_T^i \right).
\end{aligned}
\end{equation}}
\vspace{-1em}

\subsection{Training objective}
\label{objective}
We train our model by maximizing the conditional log-likelihood $\log p_\theta\left(\bm{y}_T \mid \bm{X}_{<T}\right)$. The variational evidence lower bound (\textit{ELBO}) of the model objective is derived as follows:

\vspace{-1em}
{\small
\begin{equation}
 \begin{aligned}
&\log p_\theta\left(\bm{y}_T \mid \bm{X}_{<T}\right)=\log \int_{\bm{G}}\frac{q_\phi\left(\bm{G}\right)}{q_\phi\left(\bm{G}\right)} p_\theta\left(\bm{y}_T \mid \bm{X}_{<T}, \bm{G}\right)p\left(\bm{G} \right)d\bm{G} \\
 &\geq \int_{\bm{G}} q_\phi\left(\bm{G}\right) \log p_\theta\left(\bm{y}_T \mid  \bm{X}_{<T},\bm{G} \right)p\left(\bm{G}\right)d\bm{G}
+H\left(q_\phi\left(\bm{G}\right)\right)  \\
 &\geq E_{q_\phi\left(\bm{G}\right)} [\log p_\theta\left(\bm{y}_T \mid  \bm{X}_{<T},\bm{G} \right)+ \log p\left(\bm{G}\right)]+ H\left(q_\phi\left(\bm{G}\right)\right) \\
 &\geq E_{q_\phi\left(\bm{G}\right)}\left[\sum_{i=1}^D \log p_{z_i}\left(z_T^i \right)+ \log p\left(\bm{G}\right)\right] + H\left(q_\phi\left(\bm{G}\right)\right).
\end{aligned}
\end{equation}}
\vspace{-1em}

Here, $p\left(\bm{G}\right)$ represents the prior of causal graph, and $H\left(q_\phi\left(\bm{G}\right)\right)$ is the entropy of the posterior approximator. $\log p_\theta\left(\bm{y}_T \mid \bm{X}_{<T}, \bm{G}\right)=\log p_{z}\left(\bm{z}_T \right) $ is the log-likelihood of the target distribution, in which $\bm{z}_T$ is calculated by Equation~\ref{eq:anm} at training stage.

Besides, we further adopt the binary cross entropy loss as another objective $\textit{BCE}\left(\bm{g}_T, \bm{y}_T\right)$ to improve the learning performance, where $\bm{g}_T$ is the ground truth movement at target trading day $T$. Overall, the final training loss $\mathcal{L}$ is as follows,

\vspace{-1em}
{\small
\begin{equation}
\begin{aligned}
\textit{BCE}\left(\bm{g}_T, \bm{y}_T\right)&=-\sum_{i=1}^{D}\left(g_T^{i} \log \left( y_T^i\right)+\left(1\!-\!g_T^{i}\right) \log \left(1\!-\! y_T^i\right)\right)\\
\mathcal{L} &= \frac{1}{D}\left(-\textit{ELBO}+\lambda \textit{BCE}(\bm{g}_T, \bm{y}_T)\right)
\end{aligned}
\end{equation}}
\vspace{-1em}

where $\lambda$ is the scalar weight to balance loss terms. We note that the required assumptions and the theoretical guarantees are summarized in Appendix~\ref{app:assumptions}.

\section{Experiments}
\label{experiments}

\subsection{Experimental setup}

Except for the news-driven multi-stock movement prediction task, our model could also handle the multi-stock movement prediction task without news by removing the Denoised News Encoder. Thus, we do the experiments for both two tasks. 

\textbf{Dataset} (Appendix~\ref{app:dataset}): We train and evaluate our model and baselines on six datasets:  ACL18~\cite{xu-cohen-2018-stock}, CMIN-US~\cite{luo2023causality}, CMIN-CN~\cite{luo2023causality}, KDD17~\cite{zhang2017stock}, NI225~\cite{yoo2021accurate}, and FTSE100~\cite{yoo2021accurate}. The first three of them including both historical prices and text data are used for news-driven multi-stock movement prediction task evaluation, while the last three are for multi-stock movement prediction task evaluation without news data. \textbf{Evaluation metrics}
 (Appendix ~\ref{app:metrics}): We evaluate the prediction performance of models by Accuracy (ACC) and Matthews Correlation Coefficients (MCC). \textbf{Baselines} (Appendix ~\ref{app:baselines}): HAN ~\cite{hu2018listening}, Stocknet ~\cite{xu-cohen-2018-stock},  PEN ~\cite{li2023pen}, CMIN ~\cite{luo2023causality} for news-driven multi-stock movement prediction task. LSTM ~\cite{nelson2017stock}, ALSTM ~\cite{qin2017dual}, Adv-ALSTM ~\cite{feng2018enhancing}, DTML ~\cite{yoo2021accurate} for multi-stock movement prediction task. \textbf{Parameter setup} (Appendix~\ref{app:para}): Our model is implemented with Pytorch on 4 NVIDIA Tesla V100 and optimized by Adam~\cite{kingma2014adam}. The parameter sensitivity study can be found in Appendix~\ref{app:sensitivity}.

\subsection{Results of prediction accuracy}

As shown in the top half of Table~\ref{tab:results}, CausalStock outperforms all baselines on ACC as well as MCC across three datasets demonstrating robustness performance for news-driven multi-stock movement prediction task. For the multi-stock movement prediction task, the results are reported in the bottom half of Table \ref{tab:results}. As can be seen, CausalStock exceeds all baselines across three datasets with stable performance. Overall, the results demonstrate that the proposed CausalStock can indeed improve the performance for two stock movement prediction tasks, showing the strong capabilities in handling financial texts and discovering causal relations among stocks.

\begin{table*}[t]
  \centering
  \caption{Main results of CausalStock and baselines for two stock movement prediction tasks on multiple datasets. Following the setting of baselines, the standard deviations are calculated across 10 runs for the news-driven task and 5 runs for the task without news.}
  \resizebox{\linewidth}{!}{
    \begin{tabular}{ccccccc}
    \toprule
    \multicolumn{7}{c}{\textbf{News-driven Multi-stock movement prediction task}} \\
    \midrule
    \multirow{2}[4]{*}{\textbf{Models}} & \multicolumn{2}{c}{\textbf{ACL18 (US)}} & \multicolumn{2}{c}{\textbf{CMIN-US (US)}} & \multicolumn{2}{c}{\textbf{CMIN-CN (CN)}} \\
\cmidrule{2-7}          & \textbf{ACC} & \textbf{MCC} & \textbf{ACC} & \textbf{MCC} & \textbf{ACC} & \textbf{MCC} \\
    \midrule
    \textbf{HAN} & 57.64 ± 0.0040 & 0.0518 ± 0.0050 & 53.72 ± 0.0020 & 0.0103 ± 0.0015 & 53.59 ± 0.0037 & 0.0159 ± 0.0026 \\
    \textbf{StockNet} & 58.23 ± 0.0030 & 0.0808 ± 0.0071 & 52.46 ± 0.0041 & 0.0220 ± 0.0025 & 54.53 ± 0.0062 & 0.0450 ± 0.0043 \\
    \textbf{PEN} & 59.89 ± 0.0090 & 0.1556 ± 0.0018 & 53.20 ± 0.0051 & 0.0267 ± 0.0023 & 54.83 ± 0.0086 & 0.0857 ± 0.0065 \\
    \textbf{CMIN} & 62.69 ± 0.0029 & 0.2090 ± 0.0016 & 53.43 ± 0.0085 & 0.0460 ± 0.0055 & 55.28 ± 0.0094 & 0.1110 ± 0.0990 \\
    \textbf{CausalStock} & \bf{63.42 ± 0.0039} & \bf{0.2172 ± 0.0017} & \bf{54.64 ± 0.0083} & \bf{0.0481 ± 0.0057} & \bf{56.19 ± 0.0084} & \bf{0.1417 ± 0.0813} \\
    \midrule
    \multicolumn{7}{c}
    {
    \textbf{Multi-stock movement prediction task}} \\
    \midrule
    \multirow{2}[4]{*}{\textbf{Models}} & \multicolumn{2}{c}{\textbf{KDD17 (US)}} & \multicolumn{2}{c}{\textbf{NI225 (JP)}} & \multicolumn{2}{c}{\textbf{FTSE100 (UK)}} \\
\cmidrule{2-7}          & \textbf{ACC} & \textbf{MCC} & \textbf{ACC} & \textbf{MCC} & \textbf{ACC} & \textbf{MCC} \\
    \midrule
    \textbf{LSTM} & 51.18 ± 0.0066 & 0.0187 ± 0.0110 & 50.79 ± 0.0079 & 0.0148 ± 0.0162 & 50.96 ± 0.0065 & 0.0187 ± 0.0129 \\
    \textbf{ALSTM} & 51.66 ± 0.0041 & 0.0316 ± 0.0119 & 50.60 ± 0.0066 & 0.0125 ± 0.0139 & 51.06 ± 0.0038 & 0.0231 ± 0.0077 \\
    \textbf{StockNet} & 51.93 ± 0.0001 & 0.0335 ± 0.0050 & 50.15 ± 0.0054 & 0.0050 ± 0.0118 & 50.36 ± 0.0095 & 0.0134 ± 0.0135 \\
    \textbf{Adv-ALSTM} & 51.69 ± 0.0058 & 0.0333 ± 0.0137 & 51.60 ± 0.0103 & 0.0340 ± 0.0201 & 50.66 ± 0.0067 & 0.0155 ± 0.0140 \\
    \textbf{DTML} & 53.53 ± 0.0075 & 0.0733 ± 0.0195 & 52.76 ± 0.0103 & 0.0626 ± 0.0230 & 52.08 ± 0.0121 & 0.0502 ± 0.0214 \\
    \textbf{CausalStock} & \bf{56.09 ± 0.0069} & \bf{0.1235 ± 0.0189} & \bf{53.01 ± 0.0150} & \bf{0.0640 ± 0.0310} & \bf{52.88 ± 0.0009} & \bf{0.0534 ± 0.0210} \\
    \bottomrule
    \end{tabular}}%
  \label{tab:results}%
\end{table*}%

\subsection{Ablation study}
For the ablation study, we conduct several model variants on ACL18, CMIN-CN and CMIN-US to explore the contributions of different settings in CausalStock. For the main framework, we have the following five variants. \textbf{CausalStock w/o TCD}: removing the causal discovery module from CausalStock; \textbf{CausalStock w/o News}: removing the news encoder from CausalStock and just taking prices data as input; 
\textbf{CausalStock w/o link non-existence modeling}: only model the causal link existence likelihood and leverage Sigmoid function to obtain the link existence probability;
\textbf{CausalStock w/o Lag-dependent TCD}: replacing the Lag-dependent Temporal Causal Discovery module with the Lag-independent Temporal Causal Discovery module;
\textbf{CausalStock with Variable-dependent TCD}: we add a variable-dependent causal mechanism that explicitly captures the dependencies among different stock edges. Specifically, each edge's probability is conditioned on the states of all other edges at the same time step, and the conditional function is the same as the function in the lag-dependent mechanism (Equation~\ref{eq:dependent_mechanism}). Furthermore, we explore the performance of six different Traditional News Encoders by replacing the denoised news encoder, which outputs the news embeddings as representations. \textbf{CausalStock with Glove + Bi-GRU}: leveraging the Glove word embedding and the Bi-GRU as news encoder~\cite{luo2023causality}; \textbf{CausalStock with Bert}: leveraging the pre-trained Bert (Bert-base-multilingual-cased~\cite{devlin2018bert}) as news encoder; \textbf{CausalStock with Roberta}: leveraging the pre-trained Roberta (Roberta-base~\cite{liu2019roberta}) as news encoder; \textbf{CausalStock with FinBert}: leveraging the pre-trained FinBert~\cite{araci2019finbert} as news encoder; \textbf{CausalStock with FinGPT}: leveraging the pre-trained FinGPT (FinGPT-v3.3~\cite{yang2023fingpt}) as news encoder; \textbf{CausalStock with Llama}: leveraging the pre-trained Llama ( Llama-7b-chat-hf~\cite{touvron2023llama}) as news encoder to output news embeddings.
Moreover, we explore the performance of three different LLMs for the denoised news encoder.
\textbf{CausalStock with FinGPT}: leveraging a financial LLM FinGPT (FinGPT-v3.3~\cite{yang2023fingpt}) as denoised news encoder; \textbf{CausalStock with Llama}: leveraging Llama (Llama-7b-chat-hf~\cite{touvron2023llama}) as denoised news encoder. The ablation study results are summarized in Table~\ref{tab:ablation}.

\begin{table*}[t]
  \centering
  \caption{Ablation study results on different datasets.}
 \resizebox{\linewidth}{!}{
    \begin{tabular}{cccccccc}
    \toprule
    \multirow{2}[4]{*}{\textbf{Ablation Type}} & \multirow{2}[4]{*}{\textbf{Ablation Variants}} & \multicolumn{2}{c}{\textbf{ACL18}} & \multicolumn{2}{c}{\textbf{CMIN-US}} & \multicolumn{2}{c}{\textbf{CMIN-CN}} \\
\cmidrule{3-8}         & & \textbf{ACC} & \textbf{MCC} & \textbf{ACC} & \textbf{MCC} & \textbf{ACC} & \textbf{MCC}  \\
    \midrule
    \multirow{5}{*}{Main Framework}     & \textbf{CausalStock w/o TCD} & 51.08	 & 0.0102 & 51.48 & 0.0106 & 51.37 & 0.0102  \\
    & \textbf{CausalStock w/o news} & 58.10 & 0.1421 & 53.16 & 0.0375 & 54.16 & 0.1264   \\
     & \textbf{CausalStock w/o link non-existence} & 58.21 & 0.1652	& 52.32 & 0.0241 & 53.96 & 0.0670  \\
    & \textbf{CausalStock w/o Lag-dependent TCD} & 59.19 & 0.1757 & 52.93 & 0.0312 & 54.97 & 0.1298  \\
        & \textbf{CausalStock with Variable-dependent TCD} & \textbf{63.50} & \textbf{0.2175} & 54.60 & 0.0479 & \textbf{56.25} & \textbf{0.1419}  \\
    \midrule
    \multirow{6}{*}{Traditional News Encoder} & \textbf{CausalStock with Glove+Bi-GRU} & 60.78 & 0.1952 & 53.87 & 0.0467 & 55.13 & 0.1326 \\
                         & \textbf{CausalStock with Bert} & 61.74 & 0.2067 & 53.92 & 0.0472 & 55.43 & 0.1352 \\
            
                         & \textbf{CausalStock with Roberta} & 61.81 & 0.2071 & 54.06 & 0.0477 & 55.58 & 0.1364 \\
                          & \textbf{CausalStock with FinBert} &61.72 & 0.2062	& 54.01 &0.0471 & 	55.61 & 0.1362  \\
                          & \textbf{CausalStock with FinGPT} & 61.69 & 0.2060 & 54.00 & 0.0470 & 55.60 & 0.1360 \\
                           & \textbf{CausalStock with Llama} & 62.20 & 0.2130 & 54.40 & 0.0480 & 55.85 & 0.1390  \\
    \midrule
    \multirow{3}{*}{Denoised News Encoder} & \textbf{CausalStock with FinGPT} & 61.92 & 0.2105 & 54.30 & 0.0475 & 55.67 & 0.1386 \\
                         & \textbf{CausalStock with Llama} & 62.82 & 0.2164 & 54.52 & \textbf{0.0483} & 55.97 & 0.1406 \\
                         & \textbf{CausalStock (with GPT-3.5)} & 63.42 & 0.2172 & \textbf{54.64} & 0.0481 & 56.19 & 0.1417 \\
                    
    \bottomrule
    \end{tabular}}%
  \label{tab:ablation}
\vspace{-0.5cm}
\end{table*}

We have the following observations: (1) CausalStock with news encoders all perform better than CausalStock without news, suggesting news data is particularly helpful for stock movement prediction. (2) Compared to CausalStock w/o Lag-independent TCD, CausalStock with Lag-dependent TCD has a better performance, demonstrating the value of the lag-dependent mechanism. (3) By comparing the CausalStock and CausalStock with Variable-dependent TCD, the results show that incorporating a variable-dependent causal mechanism has the potential to enhance model performance. However, the improvements are not uniform and vary depending on the dataset, which emphasizes that further validation is needed. While the above results show a promising performance of the variable-dependent causal mechanism, it significantly increases the computational complexity (from $O(L \times D^2)$ to $O(L \times D^4)$), making it challenging to apply the model to markets with large numbers of stocks. (3) By using FinGPT and Llama as the news encoder and denoised news encoder respectively, we can observe that denoised news encoders have a relatively higher ACC and MCC than their as the traditional news encoders, suggesting the value of denoised news encoders. Overall, the ablation studies show that every component contributes to CausalStock.

\subsection{Results of explainability}

Here, we present many cases detailing the interpretability of CausalStock from two perspectives: the news representation from the Denoised News Encoder, and the causal graph discovered by the Lag-dependent TCD module.

\begin{figure*}[t]
\centering
\includegraphics[width=14cm]{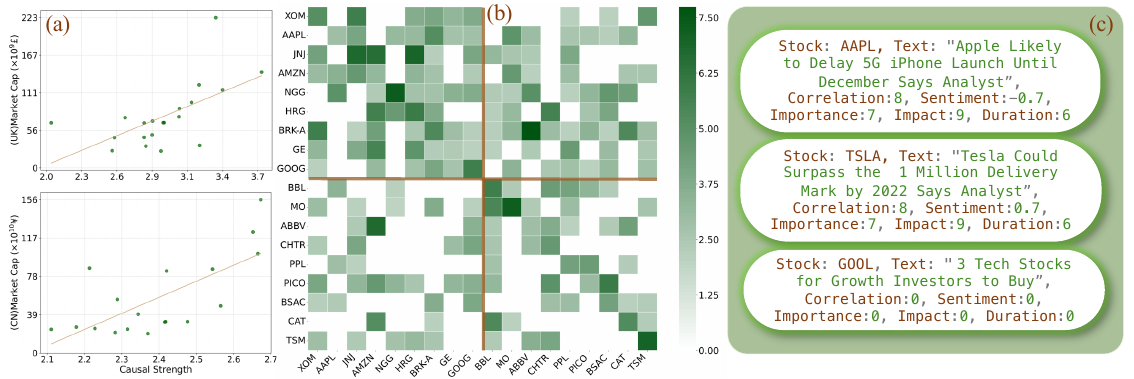}
  \caption{(a) Correlation visualization between market value and causal strength for the top 20 companies of UK and Chinese markets.
  (b) Partial causal strength matrix visualization for ACL18, encompassing the companies with the highest and lowest market values across various industries. Each matrix entry indicates causal strength between stocks, with darker shades signifying stronger causality.
  (c) Examples of denoised news encoder module output.}
\label{fig:causal}
\vspace{-0.3cm}
\end{figure*}

Firstly, regarding the Denoised News Encoder module, three cases are selected as shown in Figure \ref{fig:causal}(c).  A piece of news about APPL suggests a potential delay in its 5G iPhone launch, with Denoised News Encoder giving it a negative sentiment score of $-0.7$ and an impact score of $9$. Similarly, a news about TSLA hints at surpassing a significant delivery milestone, receiving a positive sentiment score of $0.7$. In contrast, a news piece showing no discernible connection to GOOG is scored with negligible impact. These cases indicate the Denoised News Encoder's efficacy in discerning and quantifying the potential influence of news on respective stock prices.

Secondly, concerning the causal graph discovered by Lag-dependent TCD, we denote the causal strength graph as the dot product of the causal graph $\bm{G}$ and the causal weight graph $\bm{\hat{G}}$. Every item of causal strength graph indicates not only the causality of two stocks but also the degree of causality. The visualized causal strength matrix for ACL18 is shown in Figure~\ref{fig:causal}(b) with a heatmap. From various industries, we select companies with the highest and lowest market value. The top half of Figure~\ref{fig:causal}(b) represents stocks corresponding to the nine companies with the largest market value, while the bottom half illustrates stocks from companies with the smallest. The causal strength of stocks is determined based on the average overall lags. In this heatmap, we could observe that distinct patterns emerge according to different market values. Stocks of low-market-value companies appear to have less pronounced causal relationships. We could also observe causal connections between certain high and low-market-value stocks. This is attributable to the dominant roles of large-value companies with their significant impact on those small-value firms and the stock prices.

Based on these observations, we compute the Spearman's rank correlation coefficient~\cite{spearman1961proof} between the aforementioned company's market value and their stock's causal strength on ACL18, CMIN-CN, NI225, and FSTE100 datasets, representing the US, Chinese, Japanese, and UK stock markets respectively. The correlation results are shown in Appendix~\ref{app:corr} and we also visualize some results in Figure \ref{fig:causal}(a). These results show a strong positive correlation between the market value and causal influence. This aligns with the intuition that not only do large-value companies hold pivotal economic positions, but also play crucial roles in influencing other companies. Our findings demonstrate that CausalStock does well in uncovering the causal relations within the stock market.

\subsection{Investment simulation}

\begin{wrapfigure}{R}{0.45\linewidth}
  \centering
  \vspace{-0.2cm}
  \caption{Investment simulation results.}
  \resizebox{6.5cm}{!}{
    \begin{tabular}{ccccccc}
    \toprule
    \multirow{2}[4]{*}{\textbf{Model}} & \multicolumn{2}{c}{\textbf{ACL18}} & \multicolumn{2}{c}{\textbf{KDD17}} & \multicolumn{2}{c}{\textbf{NI225}} \\
\cmidrule{2-7}          & \textbf{SR} & \textbf{APV} & \textbf{SR} & \textbf{APV} & \textbf{SR} & \textbf{APV} \\
    \midrule
    \textbf{Market Index} & 0.107 & 1.07 & 0.056 & 1.10 & 0.080 & 1.18 \\
    \textbf{PEN} & 0.293 & 1.12 & 0.132 & 1.39 & 0.171 & 1.43 \\
    \textbf{DTML} & 0.304 & 1.11 & 0.157 & 1.39 & 0.184 & 1.42 \\
    \textbf{CMIN} & 0.357 & 1.24 & 0.169 & 1.46 & 0.201 & 1.51 \\
    \textbf{CausalStock} & \textbf{0.369} & \textbf{1.32} & \textbf{0.192} & \textbf{1.49} & \textbf{0.259} & \textbf{1.52} \\
    \bottomrule
    \end{tabular}}%
  \label{tab:investment}%
\end{wrapfigure}%

Following prior works~\cite{yoo2021accurate,luo2023causality}, we evaluate CausalStock's applicability to the real world trading scenario. We conduct a portfolio strategy by choosing the top three stocks (based on predicted probabilities) with equal weight on each day of the test set and calculate the Accumulated Portfolio Value (APV) and Sharpe Ratio (SR) for evaluation. See Appendix ~\ref{app:metrics} for a metrics details. The results on three datasets are shown in Table~\ref{tab:investment}, which indicates that CausalStock achieves higher profits, and the excellent capabilities of CausalStock to balance risk with returns.

\section{Conclusion}

In this paper, we propose a novel news-driven multi-stock movement prediction framework called CausalStock.
We design a lag-dependent temporal causal discovery mechanism to uncover the causal relations among the stocks. Then the functional causal model is employed to encapsulate causal relations and predict future movements. The effectiveness of CausalStock is demonstrated by experiments on multiple real-world datasets. Moreover, CausalStock could offer a clear prediction process with explainability.

\begin{ack}
This work was supported by the National Natural Science Foundation of China (NSFC Grant No. 62122089), Beijing Outstanding Young Scientist Program NO. BJJWZYJH012019100020098, and Intelligent Social Governance Platform, Major Innovation \& Planning Interdisciplinary Platform for the ``Double-First Class'' Initiative, Renmin University of China. Shuqi Li is supported by the Fundamental Research Funds for the Central Universities, and the Research Funds of Renmin University of China (NO.297522615221).
\end{ack}

\bibliography{neurips_2024}

\begin{thebibliography}{10}

\bibitem{araci2019finbert}
D~Araci.
\newblock Finbert: Financial sentiment analysis with pre-trained language models.
\newblock {\em arXiv preprint arXiv:1908.10063}, 2019.

\bibitem{bellot2021neural}
Alexis Bellot, Kim Branson, and Mihaela van~der Schaar.
\newblock Neural graphical modelling in continuous-time: consistency guarantees and algorithms.
\newblock {\em arXiv preprint arXiv:2105.02522}, 2021.

\bibitem{blei2017variational}
David~M Blei, Alp Kucukelbir, and Jon~D McAuliffe.
\newblock Variational inference: A review for statisticians.
\newblock {\em Journal of the American statistical Association}, 112(518):859--877, 2017.

\bibitem{bubeck2023sparks}
S{\'e}bastien Bubeck, Varun Chandrasekaran, Ronen Eldan, Johannes Gehrke, Eric Horvitz, Ece Kamar, Peter Lee, Yin~Tat Lee, Yuanzhi Li, Scott Lundberg, et~al.
\newblock Sparks of artificial general intelligence: Early experiments with gpt-4.
\newblock {\em arXiv preprint arXiv:2303.12712}, 2023.

\bibitem{cheng2023causaltime}
Yuxiao Cheng, Ziqian Wang, Tingxiong Xiao, Qin Zhong, Jinli Suo, and Kunlun He.
\newblock Causaltime: Realistically generated time-series for benchmarking of causal discovery.
\newblock {\em arXiv preprint arXiv:2310.01753}, 2023.

\bibitem{devlin2018bert}
Jacob Devlin, Ming-Wei Chang, Kenton Lee, and Kristina Toutanova.
\newblock Bert: Pre-training of deep bidirectional transformers for language understanding.
\newblock {\em arXiv preprint arXiv:1810.04805}, 2018.

\bibitem{2001News}
Ilia Dichev.
\newblock News or noise?
\newblock {\em Research in Higher Education}, 42(3):237--266, 2001.

\bibitem{feng2018enhancing}
Fuli Feng, Huimin Chen, Xiangnan He, Ji~Ding, Maosong Sun, and Tat-Seng Chua.
\newblock Enhancing stock movement prediction with adversarial training.
\newblock In {\em Proceedings of the Twenty-Eighth International Joint Conference on Artificial Intelligence, {IJCAI-19}}, pages 5843--5849. International Joint Conferences on Artificial Intelligence Organization, 7 2019.

\bibitem{geffner2022deep}
Tomas Geffner, Javier Antoran, Adam Foster, Wenbo Gong, Chao Ma, Emre Kiciman, Amit Sharma, Angus Lamb, Martin Kukla, Nick Pawlowski, et~al.
\newblock Deep end-to-end causal inference.
\newblock {\em arXiv preprint arXiv:2202.02195}, 2022.

\bibitem{gerhardus2020high}
Andreas Gerhardus and Jakob Runge.
\newblock High-recall causal discovery for autocorrelated time series with latent confounders.
\newblock {\em Advances in Neural Information Processing Systems}, 33:12615--12625, 2020.

\bibitem{glorot2010understanding}
Xavier Glorot and Yoshua Bengio.
\newblock Understanding the difficulty of training deep feedforward neural networks.
\newblock In {\em Proceedings of the thirteenth international conference on artificial intelligence and statistics}, pages 249--256. JMLR Workshop and Conference Proceedings, 2010.

\bibitem{glymour2019review}
Clark Glymour, Kun Zhang, and Peter Spirtes.
\newblock Review of causal discovery methods based on graphical models.
\newblock {\em Frontiers in genetics}, 10:524, 2019.

\bibitem{DBLP:conf/iconip/GongE21}
Jiaying Gong and Hoda Eldardiry.
\newblock Multi-stage hybrid attentive networks for knowledge-driven stock movement prediction.
\newblock In Teddy Mantoro, Minho Lee, Media~Anugerah Ayu, Kok~Wai Wong, and Achmad~Nizar Hidayanto, editors, {\em Neural Information Processing - 28th International Conference, {ICONIP} 2021, Sanur, Bali, Indonesia, December 8-12, 2021, Proceedings, Part {IV}}, volume 13111 of {\em Lecture Notes in Computer Science}, pages 501--513. Springer, 2021.

\bibitem{gong2022rhino}
Wenbo Gong, Joel Jennings, Cheng Zhang, and Nick Pawlowski.
\newblock Rhino: Deep causal temporal relationship learning with history-dependent noise.
\newblock {\em arXiv preprint arXiv:2210.14706}, 2022.

\bibitem{hu2018listening}
Ziniu Hu, Weiqing Liu, Jiang Bian, Xuanzhe Liu, and Tie-Yan Liu.
\newblock Listening to chaotic whispers: A deep learning framework for news-oriented stock trend prediction.
\newblock In {\em Proceedings of the eleventh ACM international conference on web search and data mining}, pages 261--269, 2018.

\bibitem{jang2016categorical}
Eric Jang, Shixiang Gu, and Ben Poole.
\newblock Categorical reparameterization with gumbel-softmax.
\newblock {\em arXiv preprint arXiv:1611.01144}, 2016.

\bibitem{jiang2021applications}
Weiwei Jiang.
\newblock Applications of deep learning in stock market prediction: recent progress.
\newblock {\em Expert Systems with Applications}, 184:115537, 2021.

\bibitem{khemakhem2021causal}
Ilyes Khemakhem, Ricardo Monti, Robert Leech, and Aapo Hyvarinen.
\newblock Causal autoregressive flows.
\newblock In {\em International conference on artificial intelligence and statistics}, pages 3520--3528. PMLR, 2021.

\bibitem{kim2019hats}
Raehyun Kim, Chan~Ho So, Minbyul Jeong, Sanghoon Lee, Jinkyu Kim, and Jaewoo Kang.
\newblock Hats: A hierarchical graph attention network for stock movement prediction.
\newblock {\em arXiv preprint arXiv:1908.07999}, 2019.

\bibitem{kingma2014adam}
Diederik~P. Kingma and Jimmy Ba.
\newblock Adam: {A} method for stochastic optimization.
\newblock In Yoshua Bengio and Yann LeCun, editors, {\em 3rd International Conference on Learning Representations, {ICLR} 2015, San Diego, CA, USA, May 7-9, 2015, Conference Track Proceedings}, 2015.

\bibitem{li2023pen}
Shuqi Li, Weiheng Liao, Yuhan Chen, and Rui Yan.
\newblock Pen: Prediction-explanation network to forecast stock price movement with better explainability.
\newblock In {\em Proceedings of the AAAI Conference on Artificial Intelligence}, volume~37, pages 5187--5194, 2023.

\bibitem{liu2019roberta}
Yinhan Liu, Myle Ott, Naman Goyal, Jingfei Du, Mandar Joshi, Danqi Chen, Omer Levy, Mike Lewis, Luke Zettlemoyer, and Veselin Stoyanov.
\newblock Roberta: A robustly optimized bert pretraining approach.
\newblock {\em arXiv preprint arXiv:1907.11692}, 2019.

\bibitem{luo2023causality}
Di~Luo, Weiheng Liao, Shuqi Li, Xin Cheng, and Rui Yan.
\newblock Causality-guided multi-memory interaction network for multivariate stock price movement prediction.
\newblock In {\em Proceedings of the 61st Annual Meeting of the Association for Computational Linguistics (Volume 1: Long Papers)}, pages 12164--12176, 2023.

\bibitem{maddison2016concrete}
Chris~J Maddison, Andriy Mnih, and Yee~Whye Teh.
\newblock The concrete distribution: A continuous relaxation of discrete random variables.
\newblock {\em arXiv preprint arXiv:1611.00712}, 2016.

\bibitem{nelson2017stock}
David~MQ Nelson, Adriano~CM Pereira, and Renato~A De~Oliveira.
\newblock Stock market's price movement prediction with lstm neural networks.
\newblock In {\em 2017 International joint conference on neural networks (IJCNN)}, pages 1419--1426. Ieee, 2017.

\bibitem{pamfil2020dynotears}
Roxana Pamfil, Nisara Sriwattanaworachai, Shaan Desai, Philip Pilgerstorfer, Konstantinos Georgatzis, Paul Beaumont, and Bryon Aragam.
\newblock Dynotears: Structure learning from time-series data.
\newblock In {\em International Conference on Artificial Intelligence and Statistics}, pages 1595--1605. PMLR, 2020.

\bibitem{pearl2009causal}
Judea Pearl.
\newblock Causal inference in statistics: An overview.
\newblock 2009.

\bibitem{qin2017dual}
Yao Qin, Dongjin Song, Haifeng Chen, Wei Cheng, Guofei Jiang, and Garrison Cottrell.
\newblock A dual-stage attention-based recurrent neural network for time series prediction.
\newblock {\em arXiv preprint arXiv:1704.02971}, 2017.

\bibitem{roy2016solving}
Subhro Roy and Dan Roth.
\newblock Solving general arithmetic word problems.
\newblock {\em arXiv preprint arXiv:1608.01413}, 2016.

\bibitem{runge2020discovering}
Jakob Runge.
\newblock Discovering contemporaneous and lagged causal relations in autocorrelated nonlinear time series datasets.
\newblock In {\em Conference on Uncertainty in Artificial Intelligence}, pages 1388--1397. PMLR, 2020.

\bibitem{runge2019detecting}
Jakob Runge, Peer Nowack, Marlene Kretschmer, Seth Flaxman, and Dino Sejdinovic.
\newblock Detecting and quantifying causal associations in large nonlinear time series datasets.
\newblock {\em Science advances}, 5(11):eaau4996, 2019.

\bibitem{sawhney2020deep}
Ramit Sawhney, Shivam Agarwal, Arnav Wadhwa, and Rajiv~Ratn Shah.
\newblock Deep attentive learning for stock movement prediction from social media text and company correlations.
\newblock In {\em Proceedings of the 2020 Conference on Empirical Methods in Natural Language Processing (EMNLP)}, pages 8415--8426, Online, November 2020. Association for Computational Linguistics.

\bibitem{schumaker2009textual}
Robert~P Schumaker and Hsinchun Chen.
\newblock Textual analysis of stock market prediction using breaking financial news: The azfin text system.
\newblock {\em ACM Transactions on Information Systems (TOIS)}, 27(2):1--19, 2009.

\bibitem{shih2019temporal}
Shun-Yao Shih, Fan-Keng Sun, and Hung-yi Lee.
\newblock Temporal pattern attention for multivariate time series forecasting.
\newblock {\em Machine Learning}, 108:1421--1441, 2019.

\bibitem{si2013exploiting}
Jianfeng Si, Arjun Mukherjee, Bing Liu, Qing Li, Huayi Li, and Xiaotie Deng.
\newblock Exploiting topic based twitter sentiment for stock prediction.
\newblock In {\em Proceedings of the 51st Annual Meeting of the Association for Computational Linguistics (Volume 2: Short Papers)}, pages 24--29, 2013.

\bibitem{spearman1961proof}
Charles Spearman.
\newblock The proof and measurement of association between two things.
\newblock 1961.

\bibitem{2014News}
Timm~O. Sprenger, Philipp~G. Sandner, Andranik Tumasjan, and Isabell~M. Welpe.
\newblock News or noise? using twitter to identify and understand company-specific news flow.
\newblock {\em Social Science Electronic Publishing}, 41(7-8):791–830, 2014.

\bibitem{0News}
Timm~O. Sprenger and Isabell~M. Welpe.
\newblock News or noise? the stock market reaction to different types of company-specific news events.
\newblock {\em SSRN Electronic Journal}, 2001.

\bibitem{touvron2023llama}
Hugo Touvron, Thibaut Lavril, Gautier Izacard, Xavier Martinet, Marie-Anne Lachaux, Timoth{\'e}e Lacroix, Baptiste Rozi{\`e}re, Naman Goyal, Eric Hambro, Faisal Azhar, et~al.
\newblock Llama: Open and efficient foundation language models.
\newblock {\em arXiv preprint arXiv:2302.13971}, 2023.

\bibitem{wang2021coupling}
Guifeng Wang, Longbing Cao, Hongke Zhao, Qi~Liu, and Enhong Chen.
\newblock Coupling macro-sector-micro financial indicators for learning stock representations with less uncertainty.
\newblock {\em AAAI21}, pages 1--9, 2021.

\bibitem{xing2018natural}
Frank~Z Xing, Erik Cambria, and Roy~E Welsch.
\newblock Natural language based financial forecasting: a survey.
\newblock {\em Artificial Intelligence Review}, 50(1):49--73, 2018.

\bibitem{xu-cohen-2018-stock}
Yumo Xu and Shay~B. Cohen.
\newblock Stock movement prediction from tweets and historical prices.
\newblock In {\em Proceedings of the 56th Annual Meeting of the Association for Computational Linguistics (Volume 1: Long Papers)}, pages 1970--1979, Melbourne, Australia, July 2018. Association for Computational Linguistics.

\bibitem{yang2023fingpt}
Hongyang Yang, Xiao-Yang Liu, and Christina~Dan Wang.
\newblock Fingpt: Open-source financial large language models.
\newblock {\em arXiv preprint arXiv:2306.06031}, 2023.

\bibitem{yoo2021accurate}
Jaemin Yoo, Yejun Soun, Yong-chan Park, and U~Kang.
\newblock Accurate multivariate stock movement prediction via data-axis transformer with multi-level contexts.
\newblock In {\em Proceedings of the 27th ACM SIGKDD Conference on Knowledge Discovery \& Data Mining}, pages 2037--2045, 2021.

\bibitem{zhang2017stock}
Liheng Zhang, Charu Aggarwal, and Guo-Jun Qi.
\newblock Stock price prediction via discovering multi-frequency trading patterns.
\newblock In {\em Proceedings of the 23rd ACM SIGKDD international conference on knowledge discovery and data mining}, pages 2141--2149, 2017.

\bibitem{zheng2018dags}
Xun Zheng, Bryon Aragam, Pradeep~K Ravikumar, and Eric~P Xing.
\newblock Dags with no tears: Continuous optimization for structure learning.
\newblock {\em Advances in neural information processing systems}, 31, 2018.

\bibitem{zheng2020learning}
Xun Zheng, Chen Dan, Bryon Aragam, Pradeep Ravikumar, and Eric Xing.
\newblock Learning sparse nonparametric dags.
\newblock In {\em International Conference on Artificial Intelligence and Statistics}, pages 3414--3425. PMLR, 2020.

\end{thebibliography}
\bibliographystyle{plain}

\appendix
\clearpage

\section{Prompt design}
\label{app:prompt}

This structured prompt encompasses three fundamental components:

System: This section defines the role of the AI. It acts as a preliminary introduction to set the tone and context for the AI. It informs the AI that its primary role is to analyze stock-related news in various dimensions such as correlation, sentiment, importance, impact on prices, and duration of impact.

Default Prompt: This segment provides detailed instructions to the AI on how to carry out its analysis. It outlines the specific criteria and the scales on which the news should be evaluated. It also provides guidance on how to handle ambiguous or non-analyzable content and finally, it prescribes the desired output format.

Input: The final section is where the user provides the specific details about the stock, the news content, and the time of publication. It acts as the data point based on which the AI will perform its analysis as instructed in the Default Prompt.

\noindent[System]

\{As a stock trading news analyst, you are a helpful and precise assistant. Your task is to analyze the correlation between news and the given stock, sentiment polarity of the news, importance of the news, the impact of the news on stock prices, and the duration of the news impact.\}

\noindent[Default Prompt]

I need you to analyze the provided stock-related news from five dimensions:

1. Correlation between the news and the given stock: Rate the correlation on a scale of 0 to 10, where a higher score indicates a stronger correlation between the news and the given stock.

2. Sentiment polarity of the news: Rate the sentiment polarity on a scale of -1 to 1, where a value closer to -1 indicates stronger negative sentiment and a value closer to 1 indicates stronger positive sentiment.

3. Importance of the news event: Rate the importance on a scale of 0 to 10, where a higher score indicates higher importance of the news event.

4. Impact of the news on stock prices: Rate the impact on a scale of 0 to 10, where a higher score indicates a greater impact of the news on stock prices.

5. Duration of the news impact: Rate the duration on a scale of 0 to 10, where a higher score indicates a longer potential duration of the news impact.

(When you encounter a situation where analysis is not possible, please try to avoid assigning all-zero scores and instead make an effort to analyze the text content and derive scores accordingly. Only when analysis is truly impossible should you assign a score of 0 to all factors.)

(Please refrain from providing analysis and simply provide the answer according to the following format.)

Output format:

Correlation: $<$Correlation score between the news and the stock$>$

Sentiment: $<$Sentiment polarity score of the news$>$

Importance: $<$Importance score of the news event$>$

Impact: $<$Impact score of the news on stock prices$>$

Duration: $<$Duration score of the news impact$>$

\noindent[Input]

[Stock Name]: \{\textit{stock name}\}

[News Content]:\{\textit{news content}\}

[Publish Time]:\{\textit{publish time}\}

\section{Assumptions and theoretical guarantees}
\label{app:assumptions}
There are some common assumptions in causal discovery.
In this paper, we assume our model satisfies the \textit{Causal Markov Property}, \textit{Minimality and Structural Identifiability}, \textit{Correct Specification}, \textit{Causal Sufficiency} and \textit{Regularity of log likelihood}. A detailed explanation can be found in~\cite{geffner2022deep}, which explains how our model satisfies these assumptions. These assumptions guarantee the validity of the causal relations discovered by CausalStock. Considering the instability of news data, we only leverage price data $\bm{P}_{<T}$ to discover causal graph $\bm{G}$ to meet the \textit{Causal Stationary} assumption. Then we use the discovered causal graph $\bm{G}$ for aggregating both news information and price information. Technically, this could be realized by detaching the gradient from $\bm{C}_{<T}$ to $\bm{G}$.

\section{Datasets \& metrics \& baselines \& parameter setting}
\label{app:setting}
\subsection{Dataset}
\label{app:dataset}
Six datasets from various countries' stock markets are employed for conducting the experiments. The first three are used for models of fundamental analysis, which include both historical prices and text data. ACL18 ~\cite{xu-cohen-2018-stock} is a collection of data from 88 stocks in 9 industries in the US market over two years. Specifically, the price vectors after preprocessing are made up of 7 entries: date, movement percent, open price, high price, low price, close price, and volume, and the text data from Twitter are treated with tokenization and cleaning. Two CMIN datasets ~\cite{luo2023causality} are published subsequently following a similar format as ACL18. CMIN-US is collected from the US market, whereas CMIN-CN comes from 300 CSI300 stocks in the Chinese market. 

The other three datasets contain historical prices only and are applied to methods of technical analysis. KDD17 ~\cite{zhang2017stock} collects prices of 50 US stocks. \cite{feng2018enhancing} proposed to transfer the raw price vectors of KDD17 into 11 temporal features for normalizing prices and capturing the interaction between different raw price entries. With this transfer calculation, NI225, and FTSE100 ~\cite{yoo2021accurate}  record 11-feature stock prices from the US, China, Japan, and UK market respectively over the different time periods. For all datasets, the train-test set split is chronological. More detailed statistics about the datasets are presented in Table \ref{tab:dataset detail} below.

\begin{table*}[htbp]
  \centering
        \caption{Dataset Description}
  \resizebox{\textwidth}{!}{
    \begin{tabular}{ccccccccc}
    \toprule
    \multirow{2}[4]{*}{\textbf{Dataset}} & \multirow{2}[4]{*}{\textbf{Country}} & \multirow{2}[4]{*}{\textbf{Stock}} & \multicolumn{3}{c}{\textbf{Data Range}} & \multicolumn{2}{c}{\textbf{Data Resource}} & \multirow{2}[4]{*}{\textbf{Price Dim}} \\
\cmidrule{4-8}          &       &       & \multicolumn{1}{c}{\textbf{Train}} & \multicolumn{1}{c}{\textbf{Valid}} & \multicolumn{1}{c}{\textbf{Test}} & \textbf{Price} & \textbf{Text} &  \\
    \midrule
    \textbf{ACL18} \footnotemark[1] & US    & 88    & \multicolumn{1}{l}{2014/01/02-2015/08/02} & \multicolumn{1}{l}{2015/08/03-2015/09/30} & \multicolumn{1}{l}{2015/10/01-2016/01/01} & Yahoo Finance & Twitter & 7 \\
    \textbf{CMIN-US} \footnotemark[2] & US    & 110   & \multicolumn{1}{l}{2018/01/01-2021/04/30} & \multicolumn{1}{l}{2021/05/01-2021/08/31} & \multicolumn{1}{l}{2021/09/01-2021/12/31} & Yahoo Finance & Yahoo & 7 \\
    \textbf{CMIN-CN} \footnotemark[2] & CN    & 300   & \multicolumn{1}{l}{2018/01/01-2021/04/30} & \multicolumn{1}{l}{2021/05/01-2021/08/31} & \multicolumn{1}{l}{2021/09/01-2021/12/31} & Yahoo Finance & Wind  & 7 \\
    \textbf{KDD17} \footnotemark[3] & US    & 50    & \multicolumn{1}{l}{2007/01/03-2015/01/01} & \multicolumn{1}{l}{2015/01/02-2016/01/03} & \multicolumn{1}{l}{2016/01/04-2017/01/01} & Yahoo Finance & -     & 11 \\
    \textbf{NI225} \footnotemark[4] & JP    & 51    & \multicolumn{1}{l}{2016/07/01-2018/03/01} & \multicolumn{1}{l}{2018/03/02-2019/01/06} & \multicolumn{1}{l}{2019/01/07-2019/12/31}  & Yahoo Finance     & -     & 11 \\
    \textbf{FTSE100} \footnotemark[4] & UK    & 24    & \multicolumn{1}{l}{2014/01/06-2017/01/03} & \multicolumn{1}{l}{2017/01/04-2017/07/03} & \multicolumn{1}{l}{2017/07/04-2018/06/30} & -     & -     & 11 \\
    \bottomrule
    \end{tabular}}%

  \label{tab:dataset detail}%
\end{table*}%

\footnotetext[1]{https://github.com/yumoxu/stocknet-dataset}
\footnotetext[2]{https://github.com/BigRoddy/CMIN-Dataset}
\footnotetext[3]{https://github.com/fulifeng/Adv-ALSTM}
\footnotetext[4]{https://datalab.snu.ac.kr/dtml}

\subsection{Metrics}
\label{app:metrics}
Given the confusion matrix $(\begin{array}{cc} tp&fn\\fp&tn\end{array})$, where $tp,fp,tn,fn$ represent the true positives, false positives, true negatives and false negatives, we calculate ACC and MCC as follows:

{\small
\begin{equation}
ACC=\frac{tp+tn}{tp+tn+fp+gn},
\end{equation}

\begin{equation}
MCC=\frac{tp \times tn-fp \times fn}{\sqrt{(tp+fp)(fn+tp)(fn+tn)(fp+tn)}}.
\end{equation}
}

Accumulated investment portfolio value (APV) shows the accumulation of wealth over time in an intuitive form and the Sharpe Ratio (SR) is probably the most widely used metric to measure a trading strategy's return compared to its risk.
The Sharpe ratio is calculated as follows,

\begin{equation}
\mathrm{APV^t}=\prod^{t}_{i=1}(1+r^i),
\end{equation}

\begin{equation}
\mathrm{SR}=\frac{\mathbb{E}\left[\mathrm{APV}^t-R_f\right]}{\mathbb{S}\left[\mathrm{APV}^t-R_f\right]},
\end{equation}

where $r^i$ is the daily return ratio on $i$-th trading day and $R_f$ is the risk-free return.

\subsection{Baselines}
\label{app:baselines}

\paragraph{For multi-stock movement prediction}

\begin{itemize}
\item \textbf{LSTM} \cite{nelson2017stock} is an LSTM-based network that is trained with a rolling window of the last 10 months. 175 technical indicators on the characteristic of stocks and 5 features on normalized historical prices jointly form the input layer and are fed into the model.
    \item \textbf{ALSTM} \cite{qin2017dual} uses attentive LSTM in both encoder and decoder. The input attention mechanism in the encoder could extract the relevant features of stock price, whereas the temporal attention mechanism in the decoder could help learn the long-term dependencies.

    \item \textbf{Adv-LSTM} \cite{feng2018enhancing} tries to improve ALSTM through adversarial training to capture the stochastic nature of stock price and ameliorate the over-fitting. During the training process, adversarial examples are generated from latent representation and integrated with clean samples to serve as input.

    \item \textbf{DTML} \cite{yoo2021accurate} exploits the correlations between stocks in three parts: compressing the multivariate historical prices of a stock into a context vector with attentive LSTM, generating multi-level context vectors by aggregating local and global context, and capturing the correlations between stocks via transformer encoder and self-attention.
 
\end{itemize}

\paragraph{For news-driven multi-stock movement prediction}

\begin{itemize}

\item \textbf{HAN} \cite{hu2018listening}: uses attention mechanism to select useful news for stock movement prediction from chaotic online resources. The framework first applies news-level attention to find out more significant news in a date and encodes the output corpus vectors with Bi-GRU. Then, another temporal attention is applied to focus on more impactful time periods.

\item \textbf{Stocknet} \cite{xu-cohen-2018-stock}: predicts stock trend based on text and price with recurrent, continuous latent variables. The model has 3 modules, which are Market Information Encoder (MIE), Variational Movement Decoder (VMD), and Attentive Temporal Auxiliary (ATA) in sequence.

    \item \textbf{PEN} \cite{li2023pen}: a model that fuses the Bi-GRU text embedding and price inputs into Shared Representation Learning (SRL) to study their interaction. SRL also yields a Vector of Salient (VOS) that could display the importance of a piece of news and display the explainability of the model. 

    \item \textbf{CMIN} \cite{luo2023causality}: integrates causality-enhanced stock correlations and text for stock movement prediction. The approach aims to cover not only the asymmetric correlations between stocks via a newly proposed causal attention mechanism but also the multi-directional interactions between text and stock correlations. In addition, two memory networks are used for selecting the relevant information in text and stock correlation.

\end{itemize}

\subsection{Parameter setting}
\label{app:para}
Our model is implemented with Pytorch on 4 NVIDIA Tesla V100 and optimized by Adam~\cite{kingma2014adam}. All parameters of our model are initialized with Xavier Initialization~\cite{glorot2010understanding}.To better explore the model's performance, we use grid search to decide on many key hyper-parameters. The learning rate is set as $1e-5$ selected from $[1e-3, 1e-4, 1e-5, 1e-6]$. The time lag $L$ is set as $5$ selected from $[3, 5, 7, 9]$. We select the price encoder hidden size from $[4, 8, 16]$ and get the best performance with size $4$. The batch size is set as $32$. The scalar weight $\lambda$ is set to $0.01$. For the traditional news encoder, the maximum word number in one piece of news and news number in one day are set to $w=20, l=10$, respectively. The embedding size of word and news are set to $d_w=50, d_m=64$, respectively. For the Lag-dependent temporal causal discovery module, $\lambda_s=1$, $h_v$ and $h_u$ are all 1-layer MLPs. For the FCM part, the neural modules $\zeta_i$, $\ell$ and $\psi$ are all 3-layer MLPs with hidden size $332$.

\paragraph{Hyper-parameter sensitivity study}
\label{app:sensitivity}

We take a further step to analyze the main parameter sensitivity of CausalStock. We tune the key hyper-parameters learning rate $lr$, maximum time lag $L$ and loss weight $\lambda$ by grid search from this combination ${lr=1e-5, L=5, \lambda=0.01}$ while controlling other parameters. Table~\ref{tab:sensitivity} presents the results of metric ACC with different parameter settings on two tasks.

\begin{table}[htbp]
  \centering
  \caption{Hyper-parameter sensitivity study results.}
\vspace{0.5cm}
   \resizebox{0.98\linewidth}{!}{
    \begin{tabular}{ccccccccccccc}
    \toprule
    \multicolumn{1}{c}{\multirow{2}[4]{*}{\textbf{Parameters}}} & \multicolumn{4}{c}{\textbf{Learning rate $lr$}} & \multicolumn{4}{c}{\textbf{Time lag $L$}} & \multicolumn{4}{c}{\textbf{Loss weight $\lambda$}} \\
\cmidrule{2-13}          & 1e-3 & 1e-4 & \textbf{1e-5} & 1e-6 & 3     & \textbf{5}     & 7     & 9     & 0     & 0.1   & \textbf{0.01}  & 0.001 \\
    \midrule
    \textbf{ACL18 (with news)} & 62.56 & 62.34 & \textbf{63.42} & 61.58 & 61.04 & \textbf{63.42} & 63.29 & 63.15 & 58.26 & 62.35 & \textbf{63.42} & 63.45 \\
    \textbf{KDD17 (w/o news)} & 55.45 & 55.69 & \textbf{56.09} & 55.13 & 54.94 & \textbf{56.09} & 55.95 & 55.94 & 53.19 & 55.57 & \textbf{56.09} & 55.45 \\
    \bottomrule
    \end{tabular}}%
  \label{tab:sensitivity}%
\end{table}%

\section{The correlation results}
\label{app:corr}%

\begin{table}[H]
  \centering
  \vspace{-0.5cm}
   \caption{The correlation of the causal strength and the market value of companies on four datasets.}
   \vspace{0.5cm}
  \resizebox{8cm}{!}{
    \begin{tabular}{ccccc}
    \toprule
    \textbf{Statistics} & \textbf{ACL 18} & \textbf{NI225} & \textbf{CMIN-CN} & \textbf{FTSE100} \\
    \midrule
    \textbf{Spearman Corr.} & 0.7939 & 0.7212 & 0.6491 & 0.8909 \\
    \textbf{P-Value} & 0.006 & 0.0185 & 0.0036 & 0.0005 \\
    \bottomrule
    \end{tabular}}%
\label{tab:corr}
\end{table}%

\section{Limitations and future works}
\label{app:limitations}
This paper explores a method that discovers causal relations based on theoretical considerations. In the future, we could try to adopt meta-learning or incremental learning training methods to update the causal graph iteratively, i.e. explore the time-varied causal graph. While the Bernoulli distribution is suitable for determining whether a causal link exists, if we want to further explore the multi-level nature of causal relationships, more complex distributions might be needed. In the future, we could improve the model in this way.

\section{Broader impacts and safety issues}
\label{app:safe}
In this paper, we designed an LLM-based Denoised News Encoder to evaluate the news from multiple perspectives by LLMs. There exists a risk that the evaluation results of LLMs may violate human values. This safety issue needs careful consideration.

\end{document}